\documentclass[runningheads]{llncs}
\usepackage[misc]{ifsym} 
\usepackage{graphicx}
\usepackage{multirow}
\begin{document}
\title{Enhancing Transferability of Adversarial Examples with Spatial Momentum}
%
%
\author{Guoqiu Wang\inst{1} \and
Huanqian Yan\inst{1} \and
Xingxing Wei\inst{2\ {\textrm{\Letter}}}}

\institute{Beijing Key Laboratory of Digital Media (DML),
School of Computer Science and Engineering, Beihang University, Beijing, China \\
\email{\{wangguoqiu, yanhq\}@buaa.edu.cn}\\
\and
Institute of Artificial Intelligence, Hangzhou Innovation Institute, Beihang University, Beijing, China\\
\email{xxwei@buaa.edu.cn}}
\maketitle              
\begin{abstract}
Many adversarial attack methods achieve satisfactory attack success rates under the white-box setting, but they usually show poor transferability when attacking other DNN models. Momentum-based attack is one effective method to improve transferability. It integrates the momentum term into the iterative process, which can stabilize the update directions by adding the gradients' temporal correlation for each pixel. We argue that only this temporal momentum is not enough, the gradients from the spatial domain within an image, i.e. gradients from the context pixels centered on the target pixel are also important to the stabilization. For that, we propose a novel method named Spatial Momentum Iterative FGSM attack (SMI-FGSM), which introduces the mechanism of momentum accumulation from temporal domain to spatial domain by considering the context information from different regions within the image. SMI-FGSM is then integrated with temporal momentum to simultaneously stabilize the gradients' update direction from both the temporal and spatial domains.
Extensive experiments show that our method indeed further enhances adversarial transferability. It achieves the best transferability success rate for multiple mainstream undefended and defended models, which outperforms the state-of-the-art attack methods by a large margin of 10\% on average. 

\end{abstract}

\section{Introduction}
Deep neural networks (DNNs) are vulnerable to adversarial examples \cite{DBLP:journals/corr/GoodfellowSS14,szegedy2013intriguing}, which are crafted by adding imperceptible perturbations to clean images, making models output wrong predictions expected by attackers. The existence of adversarial examples has raised concerns in security-sensitive applications, e.g., self-driving cars \cite{liu2019perceptual,duan2021adversarial}, face recognition \cite{guo2021meaningful,yuan2021efficient} and video monitoring \cite{yan2021}.

\begin{figure}
\centering\includegraphics[width=0.9 \textwidth]{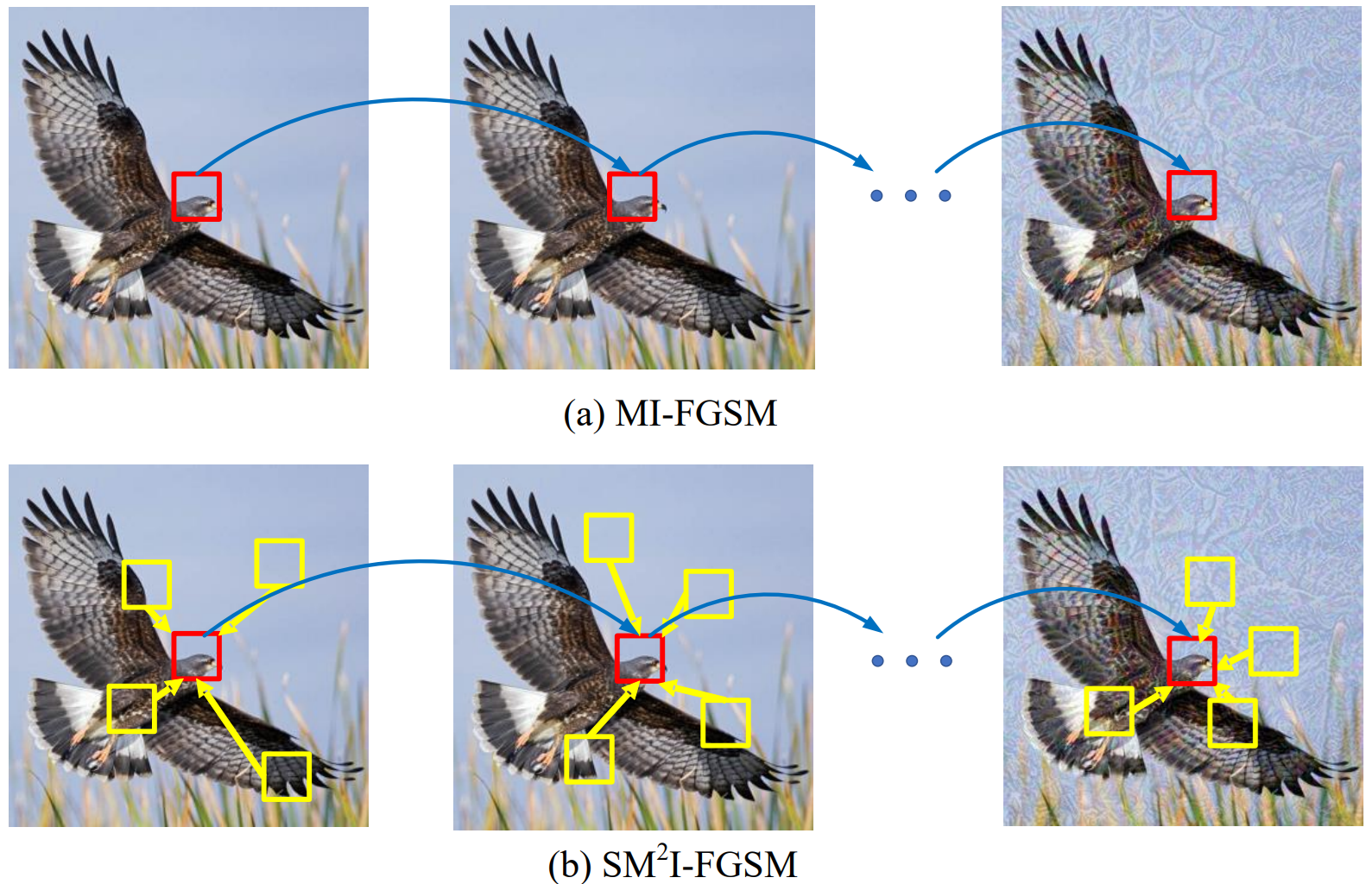}
\caption{\small Illustrations of MI-FGSM (a) and SM$^2$I-FGSM (b). MI-FGSM updates the gradient direction by accumulating the gradient in the previous iteration. SM$^2$I-FGSM considers not only the gradients’ temporal correlation but also the correlation in the spatial domain (see the yellow boxes), which can comprehensively stabilize the update direction.}
\label{fig:1}
\end{figure}

In the past years, many methods have been proposed to generate adversarial examples, such as fast gradient sign method \cite{DBLP:journals/corr/GoodfellowSS14} and its iterative variant \cite{kurakin2017adversarial}, projected gradient descent method \cite{madry2017towards} 
, and so on. They all conduct attacks in the white-box setting, utilizing the detailed information of the threat models. In addition, some works show that adversarial examples have transferability \cite{liu2016delving,papernot2017practical,wang2021improving}, which means the adversarial examples crafted for one DNN model can successfully attack other DNN models to a certain extent. The existence of transferability makes adversarial examples practical to real-world applications because attackers do not need to know the information of the target models, and thus introduces a series of serious security issues \cite{liu2019perceptual,guo2021meaningful,yuan2021efficient}.

However, most white-box attack methods usually have poor transferability. Recently, A series of methods have been proposed to address this issue, such as momentum-based iterative attack \cite{dong2018boosting,lin2019nesterov}, variance tuning iterative gradient-based method \cite{Wang_2021_CVPR},  diverse inputs \cite{xie2019improving,Wu_2021_CVPR,huang2021enhancing}, translation-invariant \cite{dong2019evading}, and multi-model ensemble attack \cite{liu2016delving}. Even so, there is still a gap between transferability attack success rates and the practical demand, which motivates us to design a more effective method to further improve adversarial transferability when attacking various DNN models.

Among the above methods to improve transferability, the momentum-based iterative attack (MI-FGSM) \cite{dong2018boosting} shows good performance and has many variants \cite{lin2019nesterov,Wang_2021_CVPR}. They integrate the momentum term into the iterative process, which can stabilize the update directions during the iterations by adding the gradients' temporal correlation to obtain better perturbations. We argue that only this temporal momentum is not enough, the gradients from the spatial domain within an image, i.e. gradients from the context pixels centered on the target pixel are also important to the stabilization. In this paper, we propose a novel method named Spatial Momentum Iterative FGSM attack, which introduces the mechanism of momentum accumulation from temporal domain to spatial domain by considering the context gradient information from different regions within the image. SMI-FGSM is then integrated with the previous MI-FGSM to construct SM$^2$I-FGSM, and thus simultaneously enhances the transferability from both the temporal and spatial domain. The attacking process is illustrated in Figure \ref{fig:1}. The main contributions can be summarized as follows:
\begin{itemize}

\item 
We show that the gradient momentum coming from the spatial domain is also useful to enhance transferability and multiple random transformations will lead to an effective spatial momentum.

\item
We propose a novel method called Spatial Momentum Iterative (SMI-FGSM) attack to improve adversarial transferability. It is then integrated with temporal momentum to simultaneously stabilize the gradients' update direction from both the temporal and spatial domains.

\item Extensive experimental results show that the proposed method could remarkably improve the attack transferability in both mainstream undefended and defended models. 
\end{itemize}

\section{Related work}
The study of adversarial attack is considered in developing robust models \cite{madry2017towards,zhao2020object}. It can be roughly categorized into two types, white-box attacks, and black-box attacks. The deep models are fully exposed to the adversary in the white-box attack setting, as their structure and parameters. Whereas in the black-box attack setting, the adversary only has little or no knowledge of the target model. Hence, black-box attacks are more practical in real-world scenarios.

Black-box attacks mainly include query-based attacks \cite{chen2017zoo,tu2019autozoom} and transfer-based attacks \cite{liu2016delving,dong2019evading}. Query-based attacks focus on estimating the gradients of the target model through interaction with the target model. However, these methods usually require a large number of queries, which is unrealistic in real-world applications. Transfer-based attacks, which are more practical and have been studied extensively, generate adversarial examples by using a white-box attack method on a source model (or source models in ensemble attack) to fool the target model. Here, we focus on improving the transfer-based black-box attacks in this paper. Some related work on adversarial attack and adversarial transferability is introduced as follows. 

Given a classification network $f_{\theta}$ parameterized by $\theta$, let $(x, y)$ denote the clean image and its corresponding ground-truth label, the goal of the adversarial attack is to find an example $x^{adv}$ which is in the vicinity of $x$ but misclassified by the network. In most cases, we use the $L_{p}$ norm to limit the adversarial perturbations below a threshold $\epsilon$, where $p$ could be $0$, $2$, $\infty$. This can be expressed as
\begin{equation}\label{eq:untarget}
f_{\theta }(x^{adv})\neq y, \ s.t. \ \left \| x^{adv}-x \right \|_{p}\leq \epsilon
\end{equation}

\textbf{Fast Gradient Sign Method (FGSM)} \cite{DBLP:journals/corr/GoodfellowSS14} generates an adversarial example $x^{adv}$ by performing one-step update as 

\begin{equation}\label{eq:fgsm}
x^{adv} = x + \epsilon \cdot sign(\bigtriangledown _{x}J(x, y))
\end{equation}
where $\bigtriangledown _{x}J$ is the gradient of the loss function $J(\cdot)$ with respect to $x$ and cross-entropy loss is often used. $sign(\cdot)$ is the sign function to limit perturbations conform to the $L_{\infty}$ norm bound. 

\textbf{The Iterative version of FGSM (I-FGSM)} \cite{kurakin2017adversarial} iteratively applies fast gradient sign method multiple times with a small step size $\alpha$, which can be expressed as
\begin{equation}\label{eq:i-fgsm}
x_{t+1}^{adv} = x_{t}^{adv} + \alpha \cdot sign(\bigtriangledown _{x_{t}^{adv}}J(x_{t}^{adv}, y))
\end{equation}

\textbf{Momentum Iterative Fast Gradient Sign Method
(MI-FGSM)} \cite{dong2018boosting} boosts the adversarial transferability by integrating the temporal momentum term into
the iterative attack to stabilize the update directions. 

\begin{equation}\label{eq:mi1}
g_{t+1} = \mu \cdot g_{t} + \frac{\bigtriangledown _{x}J(x_{t}^{adv}, y)}{||\bigtriangledown _{x}J(x_{t}^{adv}, y)||_{1}}
\end{equation}

\begin{equation}\label{eq:mi2}
x_{t+1}^{adv} = x_{t}^{adv} + \alpha \cdot sign(g_{t+1})
\end{equation}
where $g_{t}$ is the accumulated gradient and $\mu$ is the decay factor which is often set to $1.0$. MI-FGSM updates the moment gradient $g_{t+1}$ by Eq. (\ref{eq:mi1}) and then updates $x_{t+1}^{adv}$ by Eq. (\ref{eq:mi2}).

\textbf{Nesterov Iterative Fast Gradient Sign Method
(NI-FGSM)} \cite{lin2019nesterov} adapts Nesterov accelerated gradient into the iterative attacks so as to look ahead and improve the transferability of adversarial examples. NI-FGSM substitutes $x_{t}^{adv}$ in Eq. (\ref{eq:mi1}) with $x_{t}^{adv} + \alpha \cdot \mu \cdot g_{t}$.

\textbf{ Variance Tuning Momentum-based  Iterative Method
(VMI-FGSM)} \cite{Wang_2021_CVPR} further consider the gradient variance of the previous iteration to tune
the current gradient so as to stabilize the update direction. It substitutes Eq. (\ref{eq:mi1}) by 
\begin{equation}\label{eq:vmi1}
g_{t+1} = \mu \cdot g_{t} + \frac{\bigtriangledown _{x}J(x_{t}^{adv}, y) + v_{t}}{||\bigtriangledown _{x}J(x_{t}^{adv}, y) + v_{t}||_{1}}
\end{equation}
where $v_{t+1} = \frac{1}{n} \sum_{i=1}^{n}\bigtriangledown _{x}J(x_{i}, y) - \bigtriangledown _{x}J(x_{t}^{adv}, y)$, $x_{i} = x_{t}^{adv} + r_{i}$ and $r_{i}$ is the random noise within a certain range.

\textbf{Diverse Inputs (DI) attack} \cite{xie2019improving} applies random transformations to the input images at each iteration to create diverse input patterns, which brings randomness to the adversarial perturbations and improves the transferability.

\textbf{Translation-Invariant (TI) attack} \cite{dong2019evading} 
shifts the image within a small magnitude and approximately calculates the gradients by convolving the gradient of the untranslated image with a pre-defined kernel matrix. The resultant adversarial example is less sensitive to the discriminative region of the white-box model being attacked and has a higher probability to fool black-box models with defense mechanisms.

\section{Methodology}
\subsection{Spatial Momentum Iterative Attack}
In this section, we introduce the motivation and spatial momentum iterative attack in detail. We show the algorithm and its experimental results under the constraint of $L_{\infty}$ norm. This method can also be used in $L_{2}$ norm.

\begin{figure}
\centering\includegraphics[width=0.55 \textwidth]{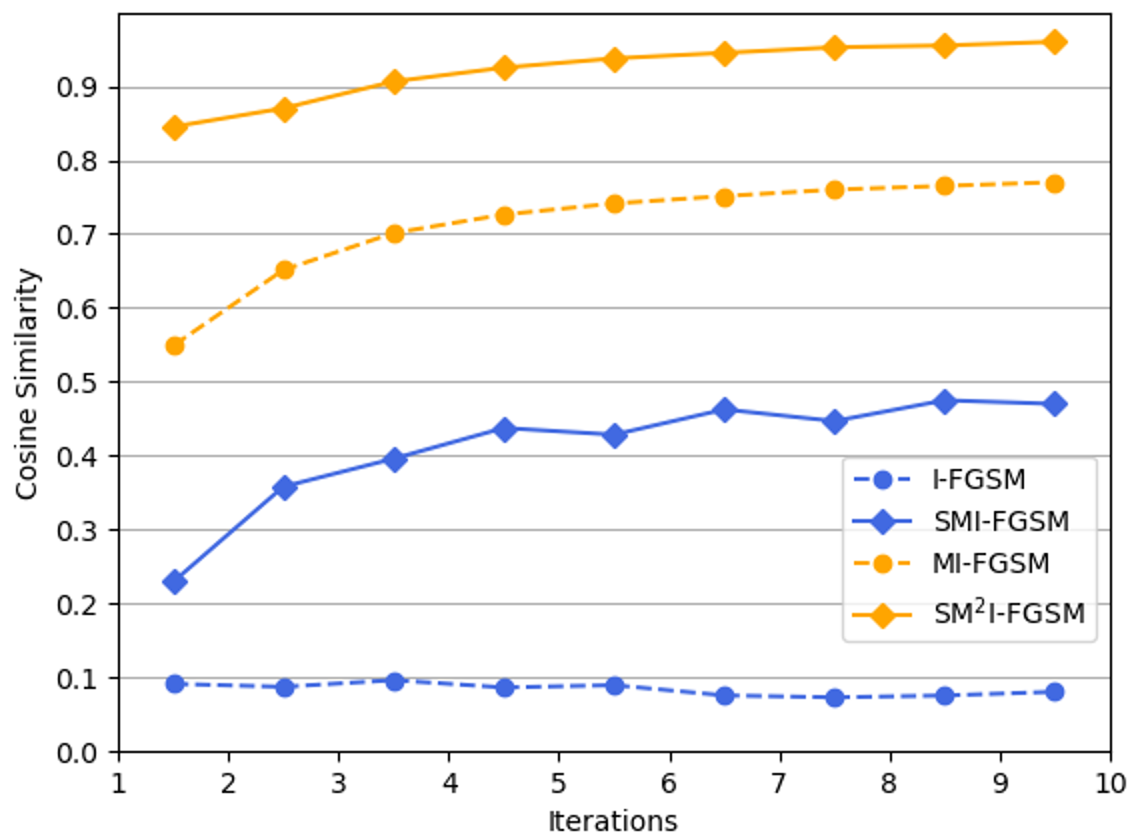}
\caption{\small The stabilization of gradients during iterations. Cosine similarity of gradients is used to measure the stabilization. The results are averaged over 1000 images. We can see that SMI-FGSM achieves better stabilization than I-FGSM and SM$^2$I-FGSM achieves the best stabilization among the four methods.}
\label{fig:cosine}
\end{figure}

MI-FGSM \cite{dong2018boosting} introduces the idea of momentum into the adversarial attack and gets a big promotion. It integrates the momentum term into the iterative process, which can be seen as adding temporal correlation (see Eq.(\ref{eq:mi1})) for the gradient that is used to update perturbations compared to I-FGSM \cite{kurakin2017adversarial} and can stabilize the update directions during the iterations (see Figure \ref{fig:cosine}). It motivates us that momentum accumulation mechanism not only can be based on the temporal domain like \cite{dong2018boosting}, but also in the spatial domain through comprehensively considering the context pixels centered on the target pixel within the image.

\begin{figure*}
\centering\includegraphics[width=0.98\textwidth]{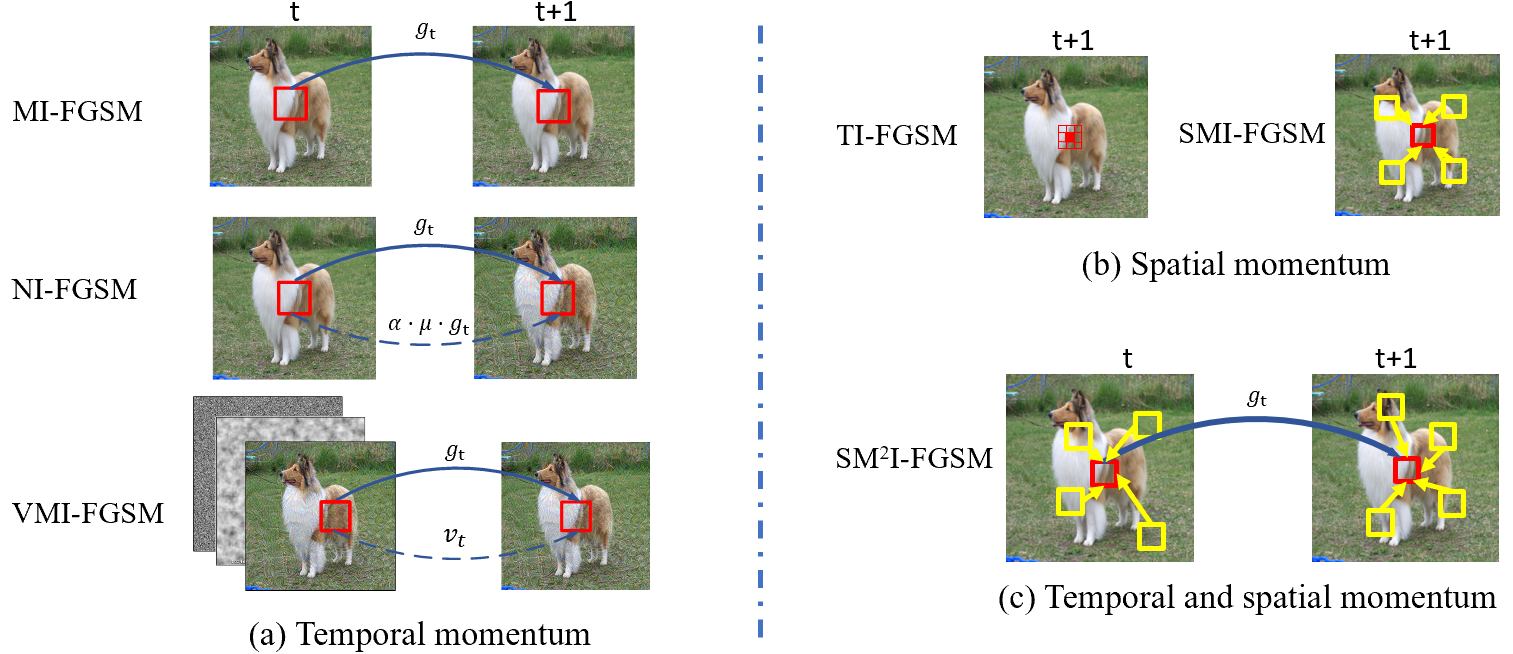}
\vspace{-0.25cm}
\caption{\small Illustrations of temporal momentum-based attacks, spatial momentum-based attacks, and their variants. There are some symbols such as $g_t$, $\alpha$, $\mu$, $v_t$, etc. Please refer to section \textit{RELATED WORK} for detail.
(a) Temporal momentum-based iterative attack.
(b) Spatial momentum-based iterative attack. 
(c) Temporal and spatial momentum-based iterative attack SM$^2$I-FGSM.}
\label{fig:3}
\end{figure*}

As Eq. (\ref{eq:i-fgsm}) shows, I-FGSM simply updates the perturbation with the gradient from image $x_{t}^{adv}$, which only considers the current pixel, while ignoring its context pixels. To stabilize the direction of updating, we propose a novel method called SMI-FGSM. It uses information from context regions by considering multiple gradients of random transformations of the same image comprehensively to generate a stable gradient.
SMI-FGSM attack is formalized as
\begin{equation}\label{eq:smi1}
g_{t+1}^{s} = \sum _{i=1}^{n} {\lambda _{i} \bigtriangledown _{x}J(H_{i}(x_{t}^{adv}), y)},  \\
x_{t+1}^{adv} = x_{t}^{adv} + \alpha \cdot sign(g_{t+1}^{s})
\end{equation}
where $H_i(\cdot)$ is used to transform $x_{t}^{adv}$ by adding random padding around the image and resizing it to its original size. The transformed image has pixel shift compared with the original image. 
$n$ denotes the number of transformation in spatial domain. $\lambda _{i}$ is the weight of $i$-th gradient, $\sum {\lambda _{i}} = 1$ and we consider $\lambda _{i} = 1/n$ in this paper. By comprehensively considering the gradients from multiple random transformations, we achieve the spatial momentum accumulation of different gradients from the context pixels.

With similar to \cite{dong2018boosting}, we use the cosine similarity of gradients during iterations to measure their similarity (Figure \ref{fig:cosine}). It can be seen that the gradients generated by considering information from different regions in the iterations have higher similarity compared to considering information from the corresponding region as I-FGSM done. This indicates the gradient generated by SMI-FGSM is more stable.

SMI-FGSM can be integrated with temporal momentum to simultaneously stabilize the gradients’ update direction from both the temporal and spatial domain (see Figure \ref{fig:cosine}) and further boost the adversarial transferability. It is named SM$^2$I-FGSM, which update $g_{t+1}$ in Eq. (\ref{eq:mi1}) by $g_{t+1}^{s}$ in Eq. (\ref{eq:smi1}).

\subsection{The Difference with Existing Attacks}
As shown in Figure \ref{fig:3}, temporal momentum-based methods stabilize the direction by using historical gradients. The basic method is MI-FGSM, which updates the current gradient using the previous gradient (illustrated by the solid line). NI-FGSM improves it by Nesterov accelerated gradient and VMI-FGSM boosts it by considering the gradient variance through adding various noises. These two methods use the previous gradient two times, which play different roles (see the solid line and dotted line in Figure \ref{fig:3}). However, they do not consider the spatial domain information, which is of the same importance as temporal domain information. The proposed SMI-FGSM considers spatial domain information during each iteration (illustrated by the yellow boxes). TI-FGSM smoothes the gradient of the untranslated image in the spatial domain by using a simple pre-defined convolution kernel and its performance is limited. 
By combining temporal and spatial momentum, SM$^2$I-FGSM can further stabilize the direction and achieves better performance.

\begin{figure}
\centering\includegraphics[width=0.98 \textwidth]{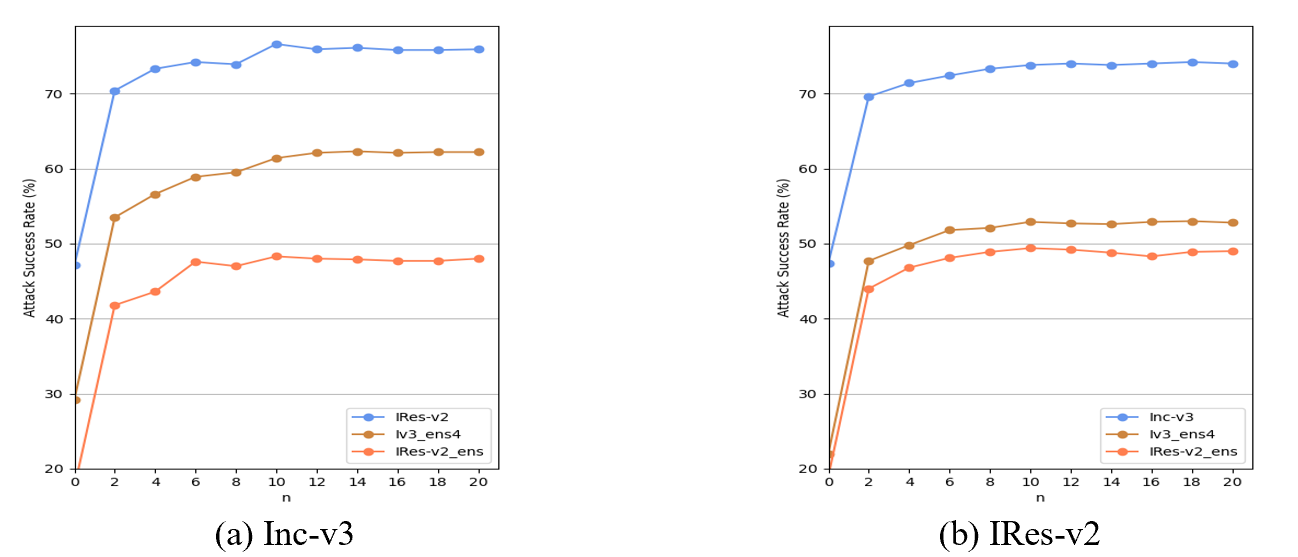}
\vspace{-0.1cm}
\caption{\small The attack success rates (\%) against three models with adversarial examples generated by SM$^2$I-FGSM on Inc-v3 (a) and Inc-v4 (b) when varying $n$.}
\label{fig:4}
\end{figure}

\section{Experiments}
\subsection{Experimental Settings}
We randomly sample 1000 images
of different categories from the ILSVRC 2012 validation
set as in \cite{dong2018boosting,Wang_2021_CVPR}. We also ensure that all of the selected images can be correctly classified by every model exploited in this work.

To evaluate our approach and compare with other mainstream methods, we  test attack performance in two normally trained models, including Inception-v3 (Inc-v3) \cite{szegedy2016rethinking}, Inception-Resnet-v2 (IRes-v2) \cite{szegedy2017inception}, and two adversarially trained models, i.e., ens4-adv-Inception-v3 (Inc-v3$_{ens4}$) and ens-adv-InceptionResNet-v2 (IRes-v2$_{ens}$) \cite{DBLP:conf/iclr/TramerKPGBM18}. In addition, three input transformation based defense strategies, including FD \cite{Liu_2019_CVPR}, BIT \cite{DBLP:conf/ndss/Xu0Q18}, and NRP \cite{naseer2020self}, are used to purify adversarial images.  After input transformations, the purified images are fed to Inc-v3$_{ens4}$  to give the final prediction.

For the settings of hyper-parameters, we follow the setting in \cite{dong2018boosting} with the maximum perturbation is 16 among all experiments with pixel values in [0, 255], the number of iteration is $10$, and step size $\alpha = 1.6$. For MI-FGSM, we set $\mu = 1.0$ as recommend in \cite{dong2018boosting}. For the transformation function $H(\cdot)$, the padding size is in [300, 330) and then resizing it to 299. We adopt
the Gaussian kernel with kernel size $5 \times 5$ for translation-invariant. And for our proposed SM$^2$I-FGSM, we set $n = 12$.

\subsection{Impact of Hyper-parameter $n$}
In SM$^2$I-FGSM, the number of $n$ plays a key role in improving the transferability. When $n$ is set to 1, SM$^2$I-FGSM will degenerate to MI-FGSM. Therefore, we practiced a series of experiments to examine the effect of $n$. We attack Inc-v3 and IRes-v2 by SM$^2$I-FGSM with different $n$ values, which range from 2 to 20 with a granularity of 2, and the results are shown in Figure \ref{fig:4}. As shown, the increase of transferability attack success rates is rapid at the beginning and then leveled off when $n$ exceeds $12$. Considering attack ability and computational complexity, $n$ is set to 12 in our experiments.

\begin{table}
\caption{The attack success rates (\%) of I-FGSM, SMI-FGSM, MI-FGSM, NI-FGSM, VMI-FGSM, and SM$^2$I-FGSM under single-model setting, $*$ indicates the white-box model being attacked. The best results are marked in bold.  We evaluate the attacks on normally trained models (i.e., Inc-v3 and IRes-v2), adversarially trained models (i.e., Inc-v3$_{ens4}$ and IRes-v2$_{ens}$), and input transformation defense strategies (i.e., FD, BIT and NRP).}
\centering  
\scalebox{1.1}{

\begin{tabular}{cccccccccc}
\hline
\multicolumn{1}{c|}{Model} & \multicolumn{1}{c|}{Attack} & \multicolumn{1}{c|}{Inc-v3} & \multicolumn{1}{c||}{IRes-v2} & \multicolumn{1}{c|}{Inc-v3$_{ens4}$} & \multicolumn{1}{c||}{IRes-v2$_{ens}$} & \multicolumn{1}{c|}{FD} & \multicolumn{1}{c|}{BIT} & \multicolumn{1}{c}{NBR} \\ \hline \vspace{-0.29cm}
 &  &  &  &  &  &  &  &  \\ \hline
\multicolumn{1}{c|}{} & \multicolumn{1}{c|}{I-FGSM} & \multicolumn{1}{c|}{99.8$^*$} & \multicolumn{1}{c||}{21.9} & \multicolumn{1}{c|}{13.2} & \multicolumn{1}{c||}{5.4} & \multicolumn{1}{c|}{12.8} & \multicolumn{1}{c|}{9.5} & \multicolumn{1}{c}{6.7} \\
\multicolumn{1}{c|}{} & \multicolumn{1}{c|}{SMI-FGSM} & \multicolumn{1}{c|}{\textbf{100.0}$^*$} & \multicolumn{1}{c||}{53.8} & \multicolumn{1}{c|}{34.2} & \multicolumn{1}{c||}{20.0} & \multicolumn{1}{c|}{29.5} & \multicolumn{1}{c|}{25.7} & \multicolumn{1}{c}{21.2} \\
\multicolumn{1}{c|}{} & \multicolumn{1}{c|}{MI-FGSM} & \multicolumn{1}{c|}{\textbf{100.0}$^*$} & \multicolumn{1}{c||}{47.9} & \multicolumn{1}{c|}{30.7} & \multicolumn{1}{c||}{18.7} & \multicolumn{1}{c|}{28.4} & \multicolumn{1}{c|}{20.1} & \multicolumn{1}{c}{16.9} \\
\multicolumn{1}{c|}{} & \multicolumn{1}{c|}{NI-FGSM} & \multicolumn{1}{c|}{\textbf{100.0}$^*$} & \multicolumn{1}{c||}{54.3} & \multicolumn{1}{c|}{34.0} & \multicolumn{1}{c||}{23.3} & \multicolumn{1}{c|}{29.4} & \multicolumn{1}{c|}{24.3} & \multicolumn{1}{c}{20.7} \\
\multicolumn{1}{c|}{} & \multicolumn{1}{c|}{VMI-FGSM} & \multicolumn{1}{c|}{\textbf{100.0}$^*$} & \multicolumn{1}{c||}{66.7} & \multicolumn{1}{c|}{47.8} & \multicolumn{1}{c||}{41.9} & \multicolumn{1}{c|}{46.6} & \multicolumn{1}{c|}{38.2} & \multicolumn{1}{c}{42.8} \\
\multicolumn{1}{c|}{\multirow{-6}{*}{Inc-v3}} & \multicolumn{1}{c|}{SM$^2$I-FGSM} & \multicolumn{1}{c|}{99.8$^*$} & \multicolumn{1}{c||}{\textbf{76.1}} & \multicolumn{1}{c|}{\textbf{61.6}} & \multicolumn{1}{c||}{\textbf{48.0}} & \multicolumn{1}{c|}{\textbf{58.9}} & \multicolumn{1}{c|}{\textbf{47.5}} & \multicolumn{1}{c}{\textbf{49.1}} \\ \hline
\multicolumn{1}{c|}{} & \multicolumn{1}{c|}{I-FGSM} & \multicolumn{1}{c|}{18.1} & \multicolumn{1}{c||}{98.6$^*$} & \multicolumn{1}{c|}{7.7} & \multicolumn{1}{c||}{4.6} & \multicolumn{1}{c|}{8.1} & \multicolumn{1}{c|}{4.3} & \multicolumn{1}{c}{5.6} \\
\multicolumn{1}{c|}{} & \multicolumn{1}{c|}{SMI-FGSM} & \multicolumn{1}{c|}{45.5} & \multicolumn{1}{c||}{97.5$^*$} & \multicolumn{1}{c|}{21.8} & \multicolumn{1}{c||}{16.3} & \multicolumn{1}{c|}{18.5} & \multicolumn{1}{c|}{16.1} & \multicolumn{1}{c}{15.6} \\
\multicolumn{1}{c|}{} & \multicolumn{1}{c|}{MI-FGSM} & \multicolumn{1}{c|}{43.6} & \multicolumn{1}{c||}{\textbf{98.8}$^*$} & \multicolumn{1}{c|}{22.2} & \multicolumn{1}{c||}{18.8} & \multicolumn{1}{c|}{19.9} & \multicolumn{1}{c|}{15.0} & \multicolumn{1}{c}{16.4} \\
\multicolumn{1}{c|}{} & \multicolumn{1}{c|}{NI-FGSM} & \multicolumn{1}{c|}{45.8} & \multicolumn{1}{c||}{97.0$^*$} & \multicolumn{1}{c|}{22.7} & \multicolumn{1}{c||}{19.5} & \multicolumn{1}{c|}{21.8} & \multicolumn{1}{c|}{18.9} & \multicolumn{1}{c}{19.3} \\
\multicolumn{1}{c|}{} & \multicolumn{1}{c|}{VMI-FGSM} & \multicolumn{1}{c|}{68.9} & \multicolumn{1}{c||}{97.2$^*$} & \multicolumn{1}{c|}{47.5} & \multicolumn{1}{c||}{42.7} & \multicolumn{1}{c|}{33.5} & \multicolumn{1}{c|}{29.8} & \multicolumn{1}{c}{31.7} \\
\multicolumn{1}{c|}{\multirow{-6}{*}{IRes-v2}} & \multicolumn{1}{c|}{SM$^2$I-FGSM} & \multicolumn{1}{c|}{\textbf{73.1}} & \multicolumn{1}{c||}{97.5$^*$} & \multicolumn{1}{c|}{\textbf{52.3}} & \multicolumn{1}{c||}{\textbf{49.8}} & \multicolumn{1}{c|}{\textbf{42.8}} & \multicolumn{1}{c|}{\textbf{36.5}} & \multicolumn{1}{c}{\textbf{40.1}} \\ \hline
\end{tabular}} \label{tab:1}
\end{table}

\vspace{-0.8cm}
\subsection{Ablation Study}
We first perform adversarial attacks using I-FGSM and SMI-FGSM under a single-model setting. The results are reported in Table \ref{tab:1}. The attacked DNN models are listed on rows, and the test DNN models are listed on columns. It is obvious that SMI-FGSM is strong as I-FGSM when attacking white-box models, they all have nearly 100\% success rate. It can be seen that the attack based on spatial momentum has significantly improved adversarial attack transferability and it is model-agnostic. For example, when we generate adversarial examples using Inc-v3 as a white-box model, SMI-FGSM and SM$^2$I-FGSM achieve success rates of 20.0\% and 48.0\% on IRes-v2$_{ens}$ respectively, while I-FGSM and MI-FGSM achieve success rates of 5.4\% and 18.7\% respectively. Through comparison and statistics, the adversarial transferability of our proposed method is ahead about 20\% than all baseline methods on average, which reveals the importance of spatial information for improving transferability.

\begin{figure}
\centering\includegraphics[width=0.9\textwidth]{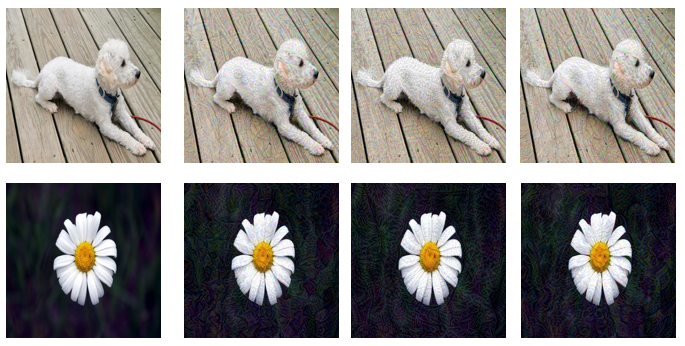}
\caption{\small Some adversarial images generated by different methods. Clean images are on the left column. The adversarial images generated by NI-FGSM, VMI-FGSM, and SM$^2$I-FGSM are on the second column, the third column, and the fourth column respectively. All the adversarial images are generated by attacking Inc-v3.}
\label{fig:5}
\end{figure}

\begin{table}[t]
\caption{The attack success rates (\%) of MI-FGSM-DTS, NI-FGSM-DTS, VMI-FGSM-DTS, and SM$^2$I-FGSM-DTS when attacking Inc-v3, $*$ indicates the white-box model being attacked. The best results are marked in bold.}
\centering  
\scalebox{1.1}{
\begin{tabular}{ccccccccc}
\hline
\multicolumn{1}{c|}{Attack} & \multicolumn{1}{c|}{Inc-v3} & \multicolumn{1}{c||}{IRes-v2} & \multicolumn{1}{c|}{Inc-v3$_{ens4}$} & \multicolumn{1}{c||}{IRes-v2$_{ens}$} & \multicolumn{1}{c|}{FD} & \multicolumn{1}{c|}{BIT} & \multicolumn{1}{c}{NBR}\\ \hline \vspace{-0.3cm}
 &  &  &  &  & \multicolumn{1}{l}{} & \multicolumn{1}{l}{} & \multicolumn{1}{l}{} &  \\ \hline
\multicolumn{1}{c|}{MI-FGSM-DTS} & \multicolumn{1}{c|}{99.7$^*$} & \multicolumn{1}{c||}{80.5} & \multicolumn{1}{c|}{70.0} & \multicolumn{1}{c||}{57.7} & \multicolumn{1}{c|}{70.4} & \multicolumn{1}{c|}{43.8} & \multicolumn{1}{c}{48.3}\\
\multicolumn{1}{c|}{NI-FGSM-DTS} & \multicolumn{1}{c|}{\textbf{99.8}$^*$} & \multicolumn{1}{c||}{82.3} & \multicolumn{1}{c|}{71.2} & \multicolumn{1}{c||}{57.9} & \multicolumn{1}{c|}{71.0} & \multicolumn{1}{c|}{44.9} & \multicolumn{1}{c}{45.5}\\
\multicolumn{1}{c|}{{VMI-FGSM-DTS}} & \multicolumn{1}{c|}{{99.6$^*$}} & \multicolumn{1}{c||}{{82.6}} & \multicolumn{1}{c|}{{77.4}} & \multicolumn{1}{c||}{{63.5}} & \multicolumn{1}{c|}{74.6} & \multicolumn{1}{c|}{50.9} & \multicolumn{1}{c}{56.4}\\
\multicolumn{1}{c|}{SM$^2$I-FGSM-DTS} & \multicolumn{1}{c|}{\textbf{99.8}$^*$} & \multicolumn{1}{c||}{\textbf{86.9}} & \multicolumn{1}{c|}{\textbf{80.7}} & \multicolumn{1}{c||}{\textbf{67.2}} & \multicolumn{1}{c|}{\textbf{79.0}} & \multicolumn{1}{c|}{\textbf{57.2}} & \multicolumn{1}{c}{\textbf{58.1}} \\ \hline
\end{tabular}} \label{tab:3}
\end{table}

\begin{table}[t]
\caption{The attack success rates (\%) of various gradient-based iterative attacks with or without DTS under the multi-model setting. The best results are marked in bold.}
\centering
\scalebox{1.1}
{
\begin{tabular}{cccccc}
\hline
\multicolumn{1}{c|}{Attack} & \multicolumn{1}{c|}{{Inc-v3$_{ens4}$ }} & \multicolumn{1}{c||}{{IRes-v2$_{ens}$ }} & \multicolumn{1}{c|}{FD} & \multicolumn{1}{c|}{BIT} & \multicolumn{1}{c}{NBR}  \\ \hline \vspace{-0.3cm}
 &  &    &  &  \\ \hline

\multicolumn{1}{c|}{MI-FGSM} & \multicolumn{1}{c|}{50.4} & \multicolumn{1}{c||}{41.8} & \multicolumn{1}{c|}{51.2} & \multicolumn{1}{c|}{36.0} & \multicolumn{1}{c}{27.9} \\
\multicolumn{1}{c|}{NI-FGSM} & \multicolumn{1}{c|}{52.9} & \multicolumn{1}{c||}{44.0} & \multicolumn{1}{c|}{51.9} & \multicolumn{1}{c|}{38.3} & \multicolumn{1}{c}{30.0} \\
\multicolumn{1}{c|}{VMI-FGSM} & \multicolumn{1}{c|}{77.4} & \multicolumn{1}{c||}{72.6} & \multicolumn{1}{c|}{68.2} & \multicolumn{1}{c|}{52.7} & \multicolumn{1}{c}{45.6} \\
\multicolumn{1}{c|}{SM$^2$I-FGSM} & \multicolumn{1}{c|}{\textbf{84.0}} & \multicolumn{1}{c||}{\textbf{79.8}} & \multicolumn{1}{c|}{\textbf{78.8}} & \multicolumn{1}{c|}{\textbf{60.4}} & \multicolumn{1}{c}{\textbf{56.3}} \\ \hline
\multicolumn{1}{c|}{MI-FGSM-DTS} & \multicolumn{1}{c|}{91.8} & \multicolumn{1}{c||}{89.4} & \multicolumn{1}{c|}{87.5} & \multicolumn{1}{c|}{70.5} & \multicolumn{1}{c}{78.8} \\
\multicolumn{1}{c|}{NI-FGSM-DTS} & \multicolumn{1}{c|}{94.5} & \multicolumn{1}{c||}{92.0} & \multicolumn{1}{c|}{88.6} & \multicolumn{1}{c|}{71.5} & \multicolumn{1}{c}{79.1} \\
\multicolumn{1}{c|}{VMI-FGSM-DTS} & \multicolumn{1}{c|}{93.4} & \multicolumn{1}{c||}{92.8} & \multicolumn{1}{c|}{89.9} & \multicolumn{1}{c|}{75.3} & \multicolumn{1}{c}{79.9} \\
\multicolumn{1}{c|}{SM$^2$I-FGSM-DTS} & \multicolumn{1}{c|}{\textbf{95.8}} & \multicolumn{1}{c||}{\textbf{94.3}} & \multicolumn{1}{c|}{\textbf{91.2}} & \multicolumn{1}{c|}{\textbf{78.4}} & \multicolumn{1}{c}{\textbf{81.3}} \\ \hline

\end{tabular}} \label{tab:4}
\end{table}

\subsection{Comparisons with State-of-the-art Attacks}
We also compare the performance of MI-FGSM and its improved versions (i.e., NI-FGSM, VMI-FGSM, SM$^2$I-FGSM) in Table \ref{tab:1}. SM$^2$I-FGSM outperforms the others by a large margin and it is model-agnostic. Particularly, if we generate adversarial images on Inc-v3, SM$^2$I-FGSM achieves an average success rate of 63.0\%, while the state-of-the-art methods NI-FGSM and VMI-FGSM achieve 40.9\% and 54.9\% respectively. We show several adversarial images generated by MI-FGSM, VMI-FGSM, and SM$^2$I-FGSM in Figure \ref{fig:5}. It can be seen that SM$^2$I-FGSM generates visually similar adversarial perturbation as others.

\subsection{Performances Combined with Other Methods}
Diverse inputs (DI), translation-invariant (TI), and scale-invariant (SI) can further improve the attack success rates individually based on I-FGSM and MI-FGSM. \cite{lin2019nesterov} has shown that the combination of them, which is called DTS in this paper, could help the gradient-based attacks achieve great transferability. We combine DTS with MI-FGSM, NI-FGSM, VMI-FGSM and SM$^2$I-FGSM as MI-FGSM-DTS, NI-FGSM-DTS, VMI-FGSM-DTS and SM$^2$I-FGSM-DTS. The results are reported in Table \ref{tab:3}. From the table, we can observe that SM$^2$I-FGSM-DTS achieves an average transferability success rate of 75.6\%. Compared to the baseline method MI-FGSM-DTS, which achieves an average transferability success rate of 67.2\%, this is a significant improvement and shows that our method has good scalability and can be combined with existing methods to further improve the success rates of transfer-based black-box attacks.

\subsection{Ensemble-based Attacks}
Related work \cite{liu2016delving} has shown that the transferability success rate can be greatly improved by using multiple models when generating adversarial examples. Here we fuse the logit outputs of different models, which is the most common ensemble method \cite{dong2018boosting}. In this subsection, we perform ensemble-based attacks by averaging the logit outputs of the models Inc-v3 and IRes-v2. The results are recorded in Table \ref{tab:4}. SM$^2$I-FGSM-DTS achieves an average attack success rate up to 88.2\% on five defense models. It is worth noting that when the transferability attack success rate exceeds 90\% in defense strategies, our method is still 2\% higher than the most advanced attack, which shows the effectiveness of the proposed method and indicates the vulnerability of current defense mechanisms. 

\section{Conclusion}
In this paper, we proposed a spatial momentum method for improving the transferability of adversarial examples, which introduces the mechanism of momentum accumulation from the temporal domain to the spatial domain. And it can be well integrated with existing attack strategies to further improve the adversarial transferability. Extensive experimental results show that the proposed method could remarkably improve the attack transferability in both excellent undefended and defended models under the single-model and multi-model settings. By comparing with the most advanced attacks, it further demonstrates the effectiveness of the proposed method. Specifically, our attack algorithm SM$^2$I-FGSM-DTS can achieve an 88.2\% transferability attack success rate on the most advanced defense strategies on average, which indicates the vulnerability of current defense mechanisms and inspire us to develop more robust models.

\vspace{+0.3cm}
\noindent
\textbf{Acknowledgements.} This work was supported by National Key R\&D Program of China (Grant No.2020AAA0104002) and the Project of the National Natural Science Foundation of China (No.62076018).

\end{document}